\documentclass{article}
\usepackage{ijcai24}

\usepackage{times}
\usepackage{soul}
\usepackage{url}
\usepackage[hidelinks]{hyperref}
\usepackage[utf8]{inputenc}
\usepackage[small]{caption}
\usepackage{graphicx}
\usepackage{amsmath}
\usepackage{amsthm}
\usepackage{booktabs}
\usepackage{algorithm}
\usepackage{algorithmic}
\usepackage[switch]{lineno}

\usepackage{amssymb}
\usepackage{pifont}
\usepackage{fontawesome}
\usepackage[table]{xcolor}
\definecolor{color0}{RGB}{211,211,211} 
\definecolor{color1}{RGB}{175,238,238} 
\definecolor{color2}{RGB}{225,255,255} 
\definecolor{color3}{RGB}{255,215,0} 
\definecolor{color4}{RGB}{255,248,220} 
\definecolor{color5}{RGB}{255,160,122} 
\definecolor{color6}{RGB}{255,228,225} 
\usepackage{multirow}
\usepackage{color}
\usepackage{hyperref}
\hypersetup{
    colorlinks=true,
    linkcolor=blue,
    filecolor=blue,      
    urlcolor=blue,
    citecolor=cyan,
}

\title{\faDiamond \ GEM: Boost Simple Network for Glass Surface Segmentation via Segment Anything Model and Data Synthesis}

\author{
Jing Hao\dag$^1$$^*$
\and
Moyun Liu\dag$^1$ 
\and
Kuo Feng Hung$^2$
\affiliations
$^1$ \ Huazhong University of Science and Technology \\
$^2$ \ The University of Hong Kong \\
\dag \ Equal Contribution. * \ Corresponding author.\\
\emails
isjinghao@gmail.com, lmomoy@hust.edu.cn, hungkfg@hku.hk
}

\begin{document}

\maketitle

\begin{abstract}
Detecting glass regions is a challenging task due to the ambiguity of their transparency and reflection properties. These transparent glasses share the visual appearance of both transmitted arbitrary background scenes and reflected objects, thus having no fixed patterns.
Recent visual foundation models, which are trained on vast amounts of data, have manifested stunning performance in terms of image perception and image generation.
To segment glass surfaces with higher accuracy, we make full use of two visual foundation models: Segment Anything (SAM) and Stable Diffusion.
Specifically, we devise a simple \textbf{G}lass surface s\textbf{E}g\textbf{M}entor named GEM, which only consists of a SAM backbone, a simple feature pyramid, a discerning query selection module, and a mask decoder.
The discerning query selection can adaptively identify glass surface features, assigning them as initialized queries in the mask decoder.
We also propose a Synthetic but photorealistic large-scale Glass Surface Detection dataset dubbed S-GSD via diffusion model with four different scales, which contain 1×, 5×, 10×, and 20× of the original real data size. This dataset is a feasible source for transfer learning. 
The scale of synthetic data has positive impacts on transfer learning, while the improvement will gradually saturate as the amount of data increases.
Extensive experiments demonstrate that GEM achieves a new state-of-the-art on the GSD-S validation set (IoU +2.1\%). Codes and datasets are available at: \href{https://github.com/isbrycee/GEM-Glass-Segmentor}{https://github.com/isbrycee/GEM-Glass-Segmentor} 
\end{abstract}

\section{Introduction}
Glass surfaces, including glass doors, windows, and walls of modern architecture, are becoming fashionable for aesthetic appeal and energy-efficient interior lighting. These glass-made items seem unfriendly towards those intelligent robots and unmanned planes operating automatically because their autonomous systems typically lack the ability to identify glass surfaces. Thus, automatic robots couldn’t avoid crashing into the glass. These transparent glasses share the same appearance with arbitrary objects behind them; that is, the glass region nearly has no fixed patterns and its boundary is ambiguous, which brings a lot of difficulty to detecting glass surfaces. Therefore, detecting the transparent obstacles required for reliable and safe robot navigation is imperative.

\begin{figure}[t!]
  \centering
  \includegraphics[width=3.35in]{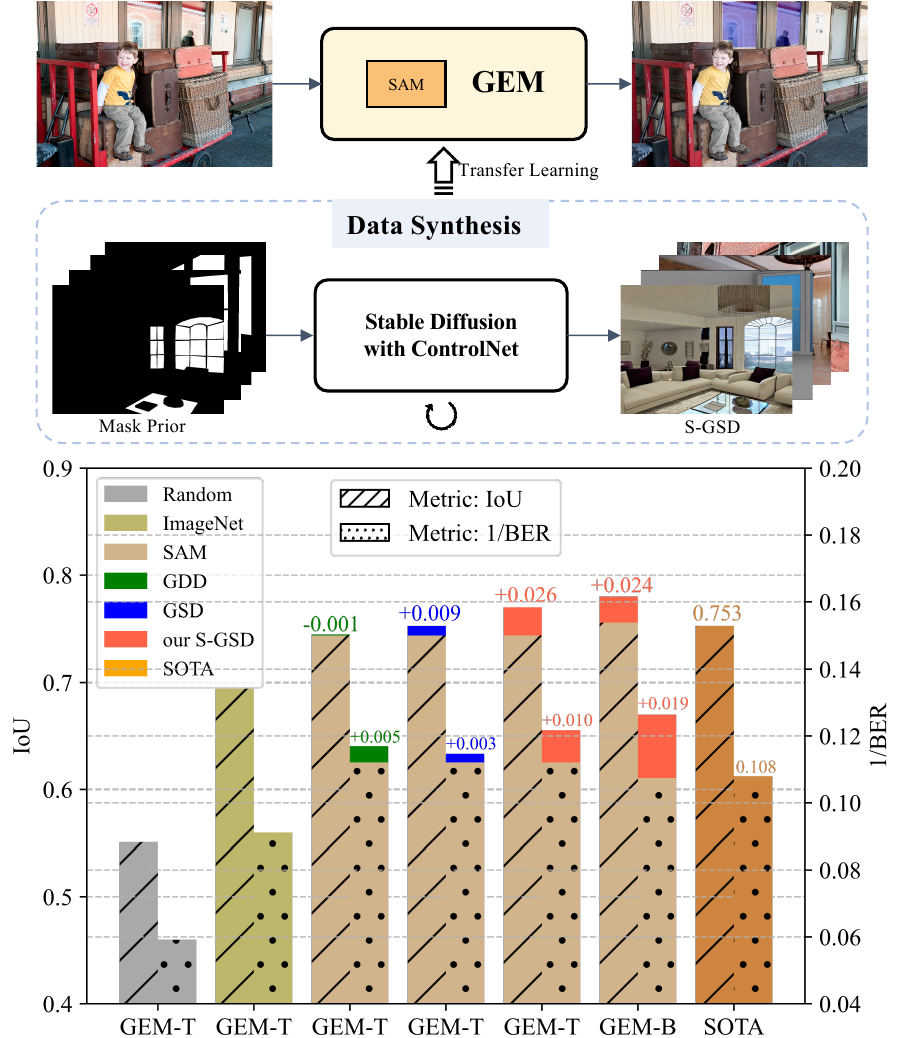}
  \caption{The upper part depicts the paradigms of glass surface segmentation with the assistance of two foundation models. The bottom part shows the comparison of the metrics IoU (Intersection over Union) and 1/BER (balance error rate) on GSD-S validation set upon different pre-trained datasets and the state-of-the-art (SOTA).  Our S-GSD dataset boosts higher improvements compared with two real glass surface datasets, GDD and GSD. The GEM-T and GEM-Base indicate the GEM-Tiny and GEM-Base, respectively. The SOTA is derived from GlassSegNet \protect\cite{lin2022exploiting}.}
  \label{fig:iou_ber}
\end{figure}
Meanwhile, AI research has witnessed a paradigm shift, with models trained on vast amounts of data at scale. ChatGPT \cite{zhang2023one} has revolutionized the NLP field, marking a breakthrough in generative AI (AIGC, a.k.a. artificial intelligence generated content). Other foundation models \cite{bommasani2021opportunities}, such as BERT, DALL-E, and DINOv2, have shown promising results in many language or vision tasks. In addition, it is evident that large-scale datasets can boost the performance of detection and segmentation tasks via transfer learning \cite{hao2023language}. Among the foundation models, Segment Anything Moel (SAM) and its variants \cite{zhang2023faster} have a distinct position as a generic image segmentation model trained on the large visual corpus. All of them have proven stunning segmentation capabilities in various natural scenarios. Besides, denoising diffusion models \cite{rombach2022high,ramesh2022hierarchical} trained at web-scale have revolutionized image generation tasks. One powerful aspect of these models is the incorporation of text-based guidance, which steers the image generation process towards an output that matches a user-provided description. This technology can generate massive samples, mitigating the problem of data scarcity to some extent. Besides, ControlNet \cite{zhang2023adding} adds spatially localized input conditions to a pretrained text-to-image diffusion model, enabling the statistical distribution of generated image to be more consistent with specific scenario tasks. With the advent of these versatile foundation models, the paradigm of traditional deep learning-based solutions, including collecting data, manually annotating, and training models, is no longer suitable. 
This leads to a natural question: {\it How can we make full use of these foundation models for serving our downstream tasks with higher accuracy and lower annotation cost?}

Utilizing these foundation models for practical application is an intriguing possibility. In this paper, we employ the SAM for the glass surface segmentation task and synthesize infinite data using the image generation model for transfer learning. By amalgamating the SAM with a simple mask decoder through some simple modifications, we design a simple \textbf{G}lass surface s\textbf{E}g\textbf{M}entor, dubbed GEM. Specifically, our GEM only consists of a ViT \cite{dosovitskiy2020image} backbone, a simple feature pyramid, a discerning query selection module, and a mask decoder. 
The framework GEM is streamlined because it does not need the contextual relationships of scenes \cite{lin2022exploiting} or object boundary prior \cite{lin2021rich} as inputs. 
We employ two different scales of feature extractors, the image encoder of SAM and MobileSAM, as the backbone for GEM.
Since GEM’s backbone is a simple plain model lacking multi-scaled features tailored for the dense downstream tasks and inspired by ViTDet \cite{li2022exploring}, we build a simple feature pyramid from a single-scale feature map for its simplicity by adding minimal adaptations upon ViT network architecture. To further address the difficulty arising from the similarity between glass and its surrounding environment, we devise a discerning query selection module. By pre-predicting the output mask through the multi-scale features, those features with high confidence scores can be used to initialize the query in the mask decoder for further refinement. It helps the decoder use prior content features from the encoder in the current image.

Additionally, we explore the impact of transfer learning on the accuracy of glass segmentation by merely using auto-generated data with precise mask annotations. 
Currently, there are only three publicly available glass datasets: GDD \cite{mei2020don}, GSD \cite{lin2021rich}, and GSD-S \cite{lin2022exploiting}, with a limited number of samples (approximately 4k samples). However, manually collecting a large-scale glass surface dataset with dense annotation is costly and time-consuming. Hence, we automatically generate a synthetic but photorealistic large-scale glass surface dataset named S-GSD with four different scales, that is, the number of images is 1x, 5x, 10x, and 20x to the real data size, respectively.
In order to construct the labeled data without any human interference, we introduce the ControlNet \cite{rombach2022high} to the Stable Diffusion \cite{rombach2022high} for controlling the spatial localization of the synthetic image. Therefore, we can collect numerous images by using the precise mask prior in the public dataset as a control condition.
These resultant image-mask pairs are used to pretrain the GEM network, as it has been proven that the pre-training \& finetune paradigm usually can boost training convergence and performance \cite{hao2023language}. A series of experiments are conducted to explore the impact of pretraining with different data scales on the performance of the glass segmentation task in terms of zero-shot learning and fine-tuning. We observed a consistent improvement in model performance for both zero-shot learning and fine-tuning when using larger synthetic data for pretraining. However, we identified a bottleneck in the improvement when the number of pretraining data becomes sufficiently large, such as being 20 times the size of the downstream training data.

We evaluate our GEM on the public dataset GSD-S, and it surpasses the previous state-of-the-art by a large margin (IoU +2.1\%). Moreover, compared with two public glass surface datasets, GDD and GSD, pretraining on S-GSD shows excellent zero-shot \& finetuning performance on GSD-S, which demonstrates the high-quality and reliability of S-GSD auto-generated by large models. Furthermore, after utilizing the pre-trained models pretrained on synthetic S-GSD, the metric IoU of GEM-Tiny and GEM-Base can be improved by 0.026 and 0.018, respectively. To the extent of our knowledge, GEM is the first work to explore how to unearth the potential of AIGC to facilitate the glass surface segmentation task. To summarize, our contributions are threefold:

\begin{itemize}
    \item Based on the foundation model, SAM, we propose a simple but accurate segmentation framework, named GEM, for glass surface segmentation. 
    \item Based on the image generation model Stable Diffusion, we automatically construct a large-scale synthesized glass surface dataset with precise mask annotation, termed S-GSD. This dataset is reliable and can be used for transfer learning. 
    \item We have conducted extensive experiments to evaluate the performance of our GEM and the quality of our S-GSD. Our method surpasses the previous state-of-the-art methods by a large margin (IoU +2.1\%). Also, our S-GSD exhibits substantial improvement on zero-shot learning and fine-tuning settings.
\end{itemize}

\begin{figure*}
\begin{center}
\centerline{\includegraphics[width=17cm]{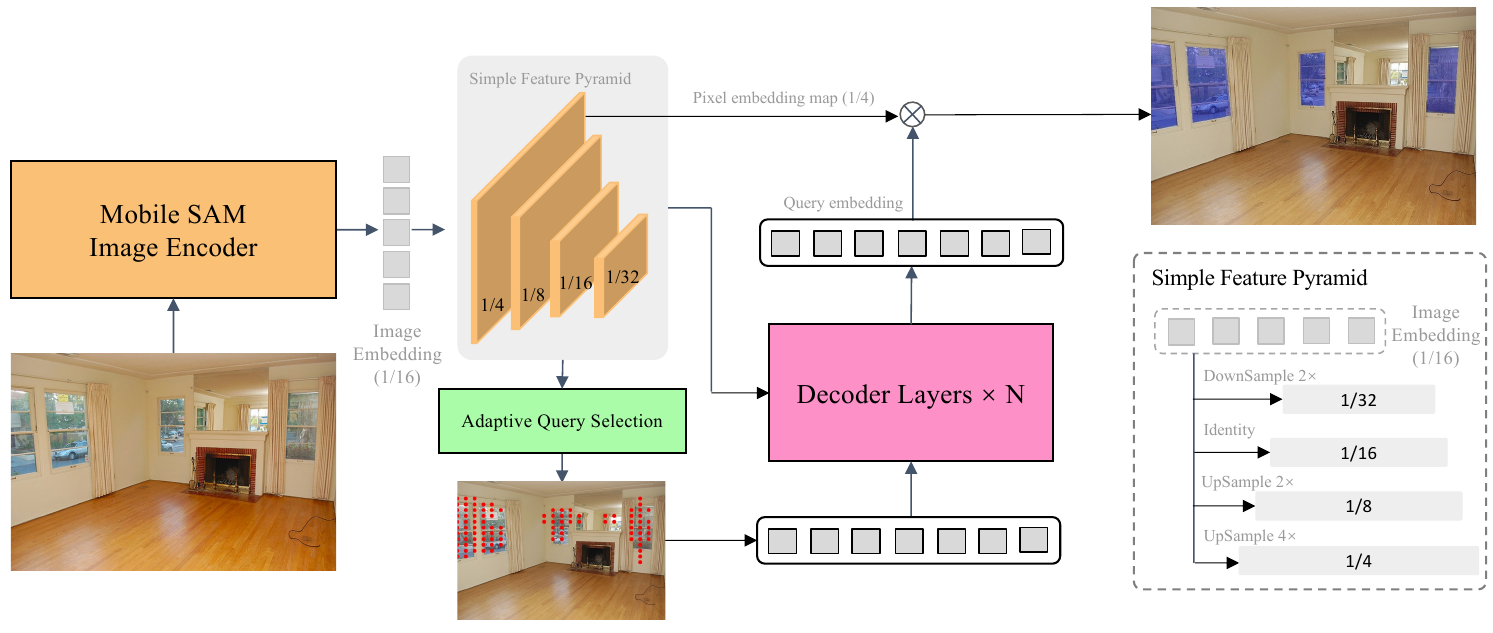}}
\end{center}
\vspace{-15pt}
\caption{The architecture of our proposed GEM. It employs a generic encoder-decoder structure, which consists of an image encoder, a simple feature pyramid, an discerning query selection, and a mask decoder. The discerning query selection is to predict the foreground and its corresponding features will be used to initialize the decoder's query.}
\label{fig:framework}
\end{figure*}

\section{Related works}

\subsection{Glass Surface Detection}
Due to the ubiquity of glass in daily scenes, accurate glass segmentation is vital for the safety of autonomous systems. Both of \cite{mei2020don} and \cite{lin2022exploiting} release real-world glass datasets and propose segmentation networks that capture rich semantic and contextual features. To enrich the representation of glass, other cues, including reflection \cite{lin2021rich}, intensity \cite{mei2022glass}, polarization \cite{mei2022glass}, or depth \cite{mei2021depth}, are used to improve the segmentation performance. However, these models heavily rely on these additional features and are not generalizable enough. Due to the transparent trait, it is hard to distinguish the glass region, so better boundary perception is exploited in \cite{han2023internal}. To obtain a more powerful model, the transformer architecture is adopted to benefit from the global attention mechanism \cite{xie2021segmenting}, which also can deal with transparency to some extent. Recently, the SAM \cite{kirillov2023segment} is also evaluated in the glass detection field \cite{han2023segment}, and it proves the visual foundation model cannot directly segment glass without any modification.

\subsection{Image Generation Foundation Model}
DPM \cite{sohl2015deep} firstly proposes a forward process and a reverse process, and it has been proven that DPM has the ability to estimate complex data distributions. Based on DPM, DPPM \cite{ho2020denoising} further successfully applies the diffusion model to the image generation task. Through the two above processes, DPPM obtains surprising generation results. To enhance the performance of the diffusion model, many works try to add extra guidance modules, such as class labels or an unconditional score estimator \cite{ho2022classifier}, which improves the image quality with auxiliary information. Instead of using classifiers, GLIDE \cite{nichol2021glide} makes the generation process conditioned on text, and the language models are exploited to boost image generation. Upainting \cite{li2022upainting} also proves that the large language model is vital to fine-grained image-text alignment. To improve the efficiency of the above two processes, Stable Diffusion \cite{rombach2022high} introduces latent space to replace pixel space. DALL-E \cite{ramesh2021zero} achieves great generalization with a scale autoregressive transformer for zero-shot text-to-image generation. Besides, DALL-E2 \cite{ramesh2022hierarchical} extends the latent space for both image and text, and the text-image latent prior is proven to make a huge contribution to the generation. To support additional input conditions for large pretrained diffusion models, ControlNet \cite{zhang2023adding} was proposed to augment models like Stable Diffusion to enable conditional inputs.

\subsection{Synthetic Data for Image Recognition}
With the advent of AIGC, generative models have been exploited to synthesize data for self-supervised pre-training and few/zero-shot learning, highlighting the transfer learning capacity of synthetic training data. He et al. \cite{he2022synthetic} have pointed out that the synthetic data delivers superior transfer learning performance for large-scale model pre-training, even outperforming ImageNet pre-training. Sariyildiz et al. \cite{sariyildiz2023fake} shows that models trained on synthetic images exhibit strong generalization properties and perform on par with models trained on real data for transfer.Marathe et al. \cite{marathe2023wedge} establishes a synthetic dataset named WEDGE that can be used to finetune state-of-the-art detectors, improving SOTA performance on real-world benchmarks. BigDatasetGAN \cite{li2022bigdatasetgan} is a synthesized ImageNet with pixel-wise annotations, and using BigDatasetGAN datasets for pretraining leads to improvements over standard ImageNet pre-training on several downstream tasks, including objection detection and segmentation. DiffuMask \cite{wu2023diffumask} uses text-guided cross-attention information to extend text-driven image synthesis to semantic mask generation. The methods trained on synthetic data in DiffuMask can achieve a competitive advantage over their counterparts in real data. In this paper, we intend to explore the labeled training data synthesis for supervised pretraining in the glass surface domain. 

\section{Method}
\subsection{Model architecture}
Our primary goal is to make full use of the knowledge of foundation models to design a simple but accurate segmentor for glass surfaces. GEM employs a generic encoder-decoder architecture that consists of an image encoder, a simple feature pyramid, discerning query selection, and a mask decoder, as shown in Fig. \ref{fig:framework}. The discerning query selection will be illustrated in Sec. \ref{discerning query selection}.

In order to harness the capabilities acquired by large models from massive corpora, we employ the image encoder derived from MobileSAM for GEM-Tiny and SAM for GEM-Base.
GEM’s encoder is a plain, non-hierarchical architecture that maintains a single-scale feature map with a scale of 1/16 throughout \cite{dosovitskiy2020image}. To boost the performance for glass surface segmentation, we use only the last feature map from the image encoder to produce multi-scale feature maps via a simple feature pyramid following ViTDet. Specifically, we generate feature maps of scales 1/8, 1/4, and 1/32 using deconvolution of strides 2 and 4 and maxpooling of strides 2, respectively.
Given these hierarchical feature maps, we simply predict masks via a mask decoder used in MaskDINO \cite{li2023mask} with simplified improvement.
The fusion operation for generating the pixel embedding map in MaskDINO is removed, and the feature map of scale 1/4 is directly appointed as the role of the pixel embedding map. We found this minimal modification did not harm the performance. Eventually, we obtain an output mask by dot-producting each content query embedding from the mask decoder with the pixel embedding map. The loss function in the training stage is followed by MaskDINO.

To summarize, an image $\mathcal{I} \in \mathbb{R}^{H\times W\times3}$ is inputted to the image encoder, and we can obtain four-scale feature maps \textit{C2}, \textit{C3}, \textit{C4}, and \textit{C5} via a simple feature pyramid $\mathcal{P}$, of which the resolutions are 1/4, 1/8, 1/16, and 1/32 of the input image, respectively.
Then the mask decoder takes the queries $\mathcal{Q} \in \mathbb{R}^{N\times256}$ and the flattened three high-level feature maps \textit{C3}, \textit{C4}, and \textit{C5} as inputs and updates queries $\mathcal{Q}$.
Finally, the updated queries $\mathcal{Q}$ dot-product with the pixel embedding map \textit{C2} to obtain a predicted mask $\mathcal{M}$. The whole process can be formulated as follows:
\begin{equation} 
\textit{C2, C3, C4, C5} = \mathcal{P}( \mathcal{E}(\mathcal{I})),
\end{equation}
\begin{equation}
M = \textit{C2} \otimes \mathcal{D}(\mathcal{Q}, Flatten(\textit{C3,C4,C5})),
\end{equation}
where $\mathcal{E}$ is the image encoder and $\mathcal{D}$ is the mask decoder. The $\otimes$ indicates the dot production. Note that the prediction masks are output at each decoder layer.

\subsection{Discerning Query Selection}
\label{discerning query selection}
\begin{figure}[t]
  \centering
  \includegraphics[width=3.35in]{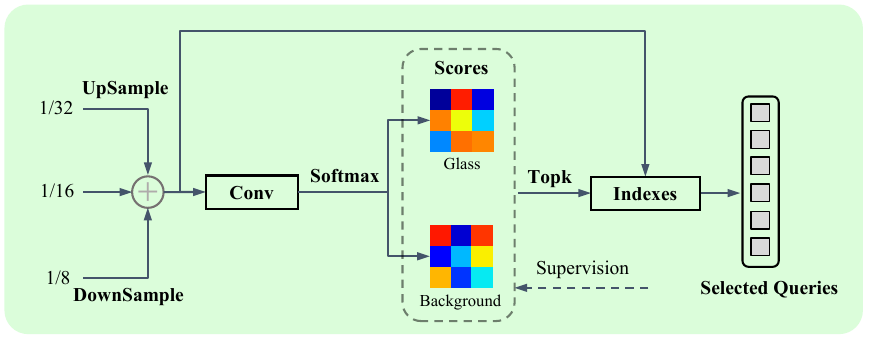}
  \caption{Illustration of the discerning query selection module.}
  \label{fig:DQS}
\end{figure}

In contrast to many common segmentation tasks, the glass regions lack distinct features and often closely resemble the background. To alleviate this challenge, we employ a discerning query selection module that is based on confidence-aware query initialization for the decoder, as shown in Fig.~\ref{fig:DQS}. Recognizing the significance of contextual information in glass segmentation \cite{lin2022exploiting}, we extend beyond relying solely on the \textit{C4} layer of the encoder for generating queries \cite{lin2022exploiting}. 
Instead, we aggregate the \textit{C3} and \textit{C5} layers with \textit{C4} layer by downsample and upsample operations, leading to the aggregated feature $\mathbf{f}$.
After that, we obtain the feature-wise classification result based on the $\mathbf{f}$ through the $Softmax$ operation.
To obtain high-quality queries, we rank all confidence scores $\mathbf{S} \in \mathbb{R}^{2hw}$ and select the features corresponding to the top-k scores as our queries.

Our discerning query selection is designed to empower the decoder by leveraging pre-classification prior content features. This selection process is effective in diverse scenarios, allowing us to adaptively identify features that significantly contribute to classification, thereby assigning them as initialized queries. This becomes particularly crucial when dealing with challenging scenarios where the color and texture of the glass are similar to those of the background, posing difficulties in discrimination. Further analysis of its functionality will be presented in Section~\ref{visualization for query selection}.

\subsection{Pre-trained Dataset Generation}
It has been proven that the paradigm of pre-training \& finetune is considerably superior to training from scratch \cite{he2022masked}, and numerous findings established that pretraining on large-scale dense prediction datasets, instead of classification datasets, could bring significant improvement for object detection and segmentation \cite{hao2023language}. However, there are only three publicly available glass datasets with a limited number of samples, and manually collecting a large-scale dataset is time-consuming and labor-intensive and may cause privacy concerns. This encourages us to ask: \textit{ How can we collect large-scale annotated pretraining data with the assistance of foundation models without any human interference?}

The advent of text-to-image generation models provides a solution for this challenge. We exploit cutting-edge generative models to automatically synthesize data for pretraining. Specifically, we utilize the ControlNet with Stable Diffusion to generate massive high-quality images by using the mask annotations in the publicly available dataset as control conditions; thus, the generated image-mask pair can construct a new segmentation training sample. In order to achieve domain customization and match the distribution between the synthetic training data and the real data, we finetune the ControlNet with real data while keeping the stable diffusion frozen. Simultaneously, we use plenty of language prompts (23 in total) to increase the diversity of synthesized images. Three examples of language prompts for image generation are as follows: 
“\textit{a photo of a clean} \textless\text{object}\textgreater”, “\textit{a close-up photo of the }\textless\text{object}\textgreater”, “\textit{a rendition of the }\textless\text{object}\textgreater”. Here, the “\textless\text{object}\textgreater” refers to \textit{transparent glasses}.

\begin{figure}[t]
  \centering
  \includegraphics[width=3.35in]{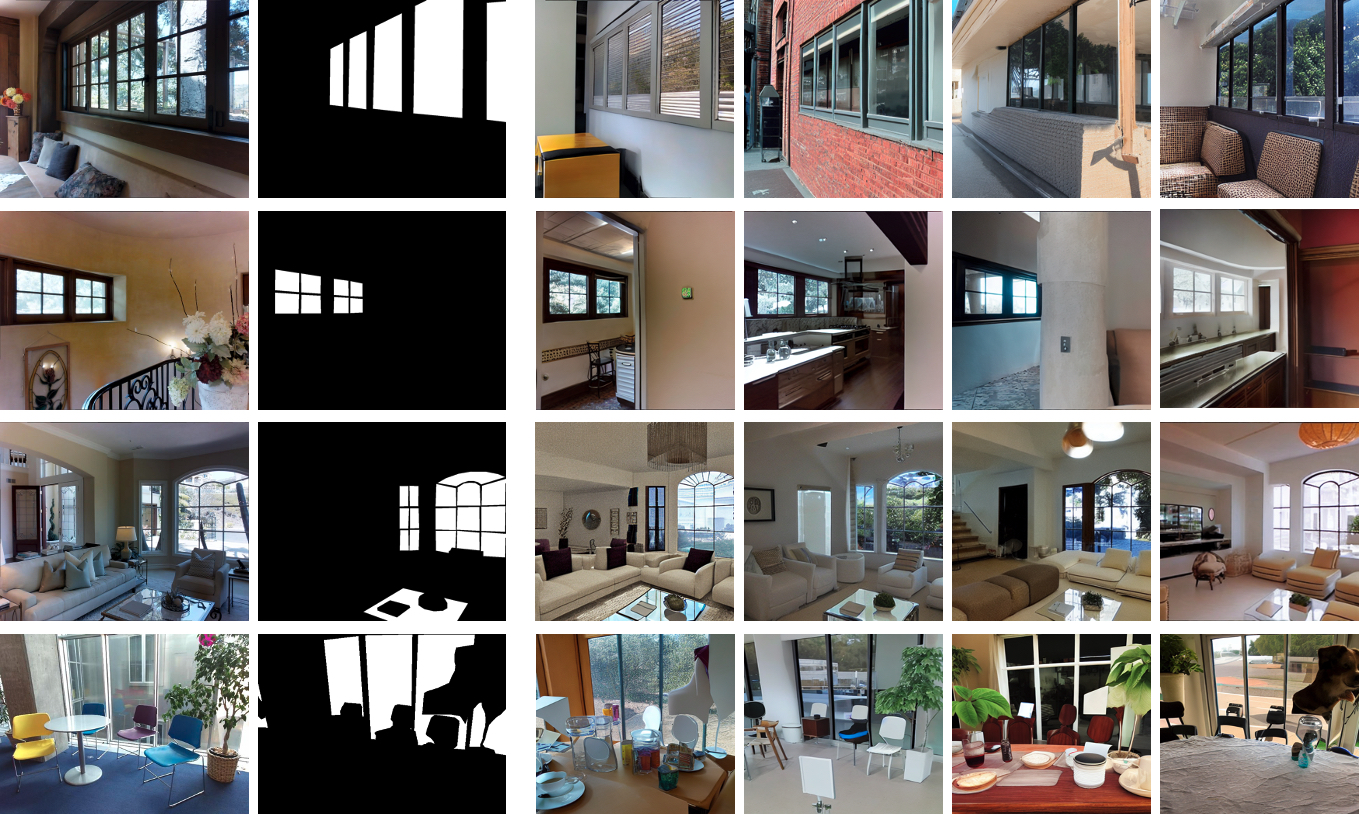}
  \caption{Visual examples of synthetic data. The first and second columns refer to the real data and corresponding mask, respectively. The rest of the columns are the synthetic data.}
  \label{fig:synthetic_data}
\end{figure}

Hence, the generation process requires the control inputs of the segmentation mask and the language prompt. Note that we don’t use the validation mask during the process of finetuning the ControlNet and synthetic data generation to avoid the leakage of validation data.
Fig. \ref{fig:synthetic_data} shows some synthetic images in several mask control conditions. Given a specific mask condition, we can generate high-quality images where the position of the glasses is strictly followes the segmentation map layout while the background is full of diversity. Additional synthetic data images and an in-depth comparison between our synthetic data and the real data can be found in the supplementary materials.

Surprisingly, the synthesized data exhibits three prominent advantages when serving as pretraining data: (1) The generated glass possesses natural properties, including transparency, reflection, and refraction, akin to real glass, adhering to the principles of physical optics. (2) The glass in the generated images is precisely aligned with the mask control condition, resulting in images that possess accurately annotated segmentation maps. These generated images and corresponding maps can be constructed as image-mask pairs for supervised pretraining. (3) The generated images encompass a diverse array of background contents and conform to the distribution observed in real images, thus providing a more rich and varied visual representation for pretraining models.

\setlength{\belowcaptionskip}{-0.2cm}
Furthermore, we investigate the impact of the scale of the generated data on zero-shot and fine-tuning performance. We generated Synthetic Glass Surface Dataset, termed S-GSD, that are 1x, 5x, 10x, and 20x larger than the training dataset GSD-S. Pretraining and fine-tuning were conducted on the GEM-Tiny and GEM-Base network architectures, respectively. Our findings indicate that as we progressively utilize larger-scale synthetic datasets for pretraining models, both zero-shot and fine-tuning results exhibit steady improvements. The detailed experimental procedures will be illustrated in Section \ref{The impact of synthetic data scale}.

\section{Experiments}
\subsection{Experimental Setup}
\noindent
\textbf{Datasets.}
We design and conduct extensive experiments to verify the performance of GEM and the effectiveness of the S-GSD dataset. We select the newest and large-scale glass surface segmentation dataset, GSD-S \cite{lin2022exploiting}, to evaluate our GEM’s performance. This dataset consists of 3911 training samples and 608 test samples. Besides, we compare our proposed S-GSD dataset with two real glass datasets, GSD and GDD, via transfer learning, which demonstrates that our auto-generated glass dataset can express similar improvements compared with real glass datasets annotated manually.

\noindent
\textbf{Implementation Details.} For fair comparison, we keep the evaluation metric and image pre-processing the same with GlassSemNet \cite{lin2022exploiting}. 
Following \cite{han2023internal}, we don't use conditional random fields (CRF) as a post-processing. The models are trained 80 epochs in total on 4 NVIDIA GeForce RTX 3060 GPUs. We utilize the ViT-Tiny in MobileSAM \cite{zhang2023faster} and the ViT-Base in SAM \cite{kirillov2023segment} as our backbone's pre-trained models. The detailed training parameters are provided in the supplementary materials.

\subsection{Comparisons with the State-of-the-Arts}
We present a comprehensive evaluation of GEM against a diverse set of state-of-the-art semantic segmentation methods, encompassing mainstream transformer-based models like SETR, Segmenter, Swin, SegFormer, Mask2Former, Mask DINO, and FASeg, as well as specialized models including GDNet, GlassNet, and GlassSemNet.
Most methods are re-trained on GSD-S, following the default training settings specified in the original papers.
The results presented in Table \ref{tab:GSD-S-Comparisons} highlight the effectiveness of GEM across four evaluation metrics. GEM outperforms all competing models, showcasing its superiority in glass surface segmentation. Notably, when compared to the GlassSemNet, which was the previous state-of-the-art method on the GSD-S dataset, both GEM-Tiny and GEM-Base achieve substantial improvements, with a notable 1.7\% and 2.1\% enhancement in the metric IoU, respectively, along with improvements across other metrics. 
Besides, we conduct comparisons of the FPS (Frame Per Second) across different methods. GEM-Tiny and GEM-Base exhibited processing speeds three times and two times faster, respectively, than the GlassSemNet, validating the efficiency of our model. Additional FPS tests for various comparative methods are presented in the supplementary materials.

In Fig.~\ref{fig:qualitative_comparison}, we present qualitative comparisons between our proposed method and other state-of-the-art networks. We highlight segmentation results in four distinct scenarios, showing the effectiveness of our GEM-Tiny alongside three other methods. Notably, our approach excels in distinguishing glass regions, outperforming other methods that often struggle, particularly when faced with similarities in color and texture features between glasses and the background. We also conducted additional tests on real-world outdoor scenes for robustness in terms of generalization, which are discussed in the supplementary materials.

\begin{table}[]
\centering
\tabcolsep=0.04cm
\caption{The evaluation results on GSD-S. Some scores of competing methods are taken from GlassSemNet.}
\begin{tabular}{>{\small}p{4.3cm} >{\centering\arraybackslash}p{1.0cm} >{\centering\arraybackslash}p{1.0cm} >{\centering\arraybackslash}p{1.0cm} >{\centering\arraybackslash}p{1.0cm}}
\toprule
Methods      & IoU$\uparrow$ & $F_{\beta}\uparrow$ &MAE$\downarrow$  &BER$\downarrow$   \\
\midrule
SCA-SOD \cite{siris2021scene}          & 0.558  & 0.689     & 0.087 & 15.03  \\
SETR \cite{zheng2021rethinking}              & 0.567  & 0.679     & 0.086 & 13.25  \\
Segmenter \cite{strudel2021segmenter}        & 0.536  & 0.645     & 0.101 & 14.02  \\
Swin \cite{liu2021swin}               & 0.596  & 0.702     & 0.082 & 11.34  \\
ViT \cite{yuan2021tokens}                & 0.562  & 0.693     & 0.087 & 14.72  \\
Twins \cite{chu2021twins}        & 0.590  & 0.703     & 0.084 & 12.43  \\
SegFormer \cite{xie2021segformer}     & 0.547  & 0.683     & 0.094 & 15.15  \\
MaskFormer \cite{cheng2021maskformer}          & 0.707  & 0.826    & 0.043 & 10.91  \\
Mask2Former\cite{cheng2021mask2former}         & 0.732  & 0.838     & 0.043 & 8.93  \\
Mask DINO \cite{li2023mask}     & 0.687   &  0.816     & 0.049  & 11.67   \\
FASeg \cite{he2023dynamic}       & 0.725  & 0.843     & 0.048 & 10.26  \\ 
MP-Former \cite{zhang2023mp}           & 0.734  & 0.827     & 0.042 & 8.67  \\
NAT \cite{hassani2023neighborhood}               & 0.730  & 0.846     & 0.041 & 10.16  \\
\midrule
GDNet \cite{mei2020don}           & 0.529  & 0.642     & 0.101 & 18.17  \\
GlassNet \cite{lin2021rich}         & 0.721  & 0.821     & 0.061 & 10.02  \\
GlassSemNet \cite{lin2022exploiting}   & 0.753  & 0.860     & 0.035 & 9.26   \\
\midrule
GEM-Tiny         & 0.770 & 0.865 & 0.032 & \textbf{8.21} \\
GEM-Base         & \textbf{0.774} & \textbf{0.865} & \textbf{0.029} & 8.35 \\
\bottomrule
\end{tabular}
\label{tab:GSD-S-Comparisons}
\end{table}

\begin{table}[]
\centering
\tabcolsep=0.15cm
\caption{The results of zero-shot \& finetune on GSD-S validation set compared with three different pretraining datasets.}
\begin{tabular}{lcccccc}
\toprule
Paradigm & Dataset  & IoU $\uparrow$    & $F_{\beta} \uparrow$ & MAE $\downarrow$  & BER $\downarrow$   \\
\midrule
             &  GDD       & 0.481 & 0.620 & 0.179 & 20.42     \\
Zero-Shot    &  GSD       & \textbf{0.714} & 0.811 & \textbf{0.054} & \textbf{10.30} \\
             &  S-GSD-1x  & 0.703 & \textbf{0.819} & 0.215 & 10.79    \\
\midrule
             &  GDD       & 0.743 & 0.843 & 0.039 &  8.55    \\
Finetune     &  GSD       & 0.753 & 0.851 & \textbf{0.037} & 8.72     \\
             &  S-GSD-1x  & \textbf{0.755} & \textbf{0.852} & 0.038 & \textbf{8.39}    \\
\bottomrule
\end{tabular}
\label{tab:Effeciveness_of_S-GSD}
\end{table}

\begin{figure}[t]
  \centering
  \includegraphics[width=3.35in]{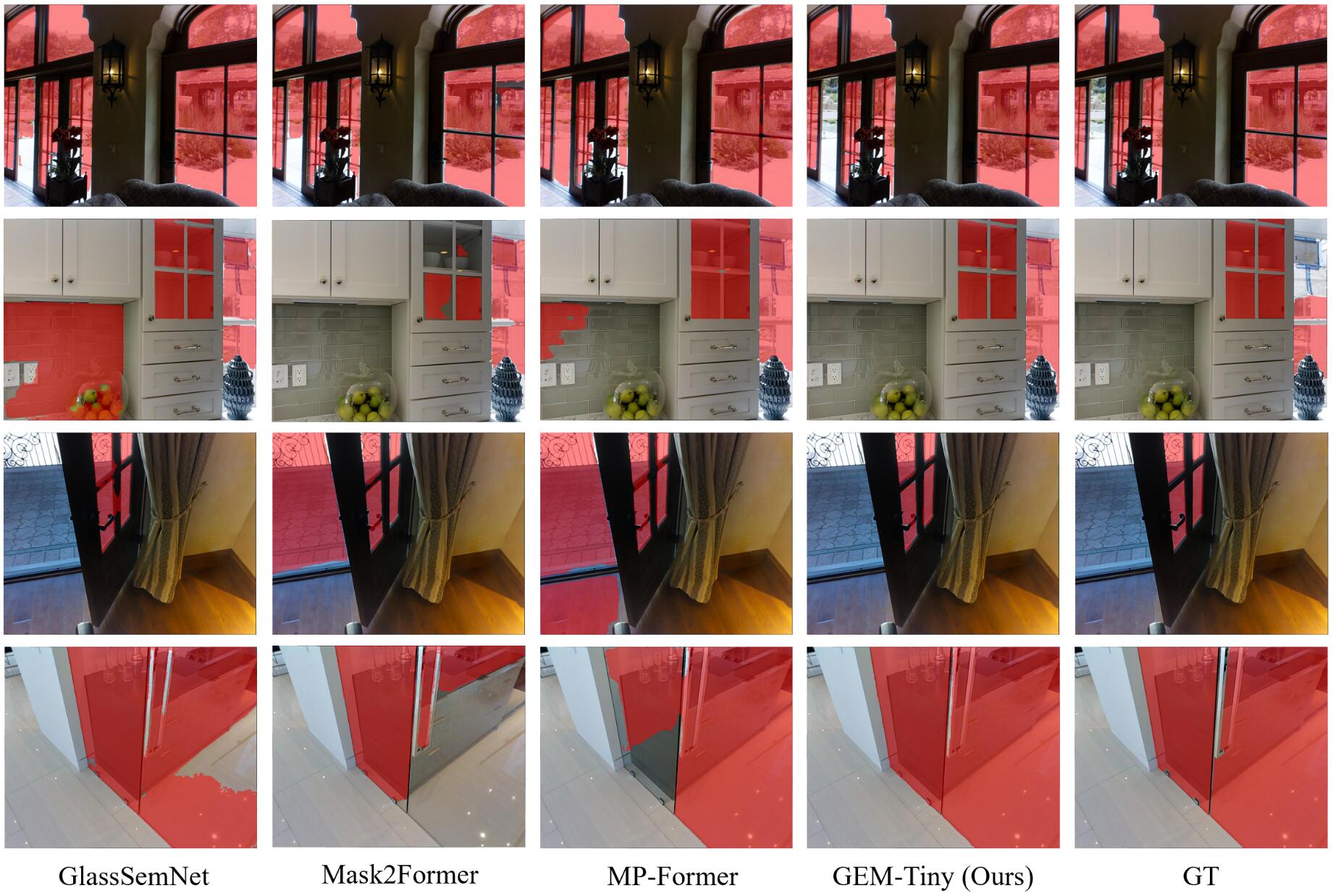}
  \caption{Qualitative comparisons of our GEM with other methods.}
  \label{fig:qualitative_comparison}
\end{figure}

\subsection{Effectiveness of auto-generated S-GSD}
To verify the quality and effectiveness of the generated S-GSD, we compared it with two real high-quality public-available datasets, GDD and GSD, in terms of the performance of zero-shot and finetune on GSD-S with the GEM-Tiny model. The number of images in GDD, GSD, and S-GSD-1x is 3916, 4102, and 3912, respectively, so the experimental results are fairly comparable. 

As for the zero-shot paradigm, the performance of GSD and S-GSD-1x exhibits an obvious increase compared with the GDD, while the metrics of GSD and S-GSD-1x are quite similar. Regarding the finetuning on GSD-S, our auto-generated pretraining dataset S-GSD-1x outperforms other real public datasets in three metrics, including IoU, $F_{\beta}$, and BER. Additionally, the metric MAE on S-GSD-1x and GSD are numerically close (0.038 vs. 0.037). These results prove that our S-GSD-1x dataset is as photorealistic as the real data, and it is indeed suitable to be used as the pretraining data. 

\subsection{The impact of synthetic data scale}
\label{The impact of synthetic data scale}
With the increase in pretraining data scale, pretrained models acquire more comprehensive visual semantic representations, leading to enhanced performance in downstream tasks after fine-tuning \cite{hao2023language}. We explore the impact of synthetic data scale on glass surface segmentation. Specifically, we generate four pretraining datasets that are 1x, 5x, 10x, and 20x larger than the GSD-S, respectively. These pretraining datasets are used to pretrain and finetune our GEM model. We pretrain 160 epochs in the pretraining stage and 80 epochs in the finetuning stage. Concerning the zero-shot for GEM-Tiny, the IoU and $F_{\beta}$ increase by 0.023 and 0.025 when the data scale expands from 1x to 20x, respectively. The same trend can be observed in the performance of finetune for GEM-Tiny, which increased by 0.015 and 0.013, respectively. As for the GEM-Base, it is also clear that the performance of zero-shot and finetune is steadily improving when the scale of pretraining data increases continuously. However, when the data scale expands from 10x to 20x, the improvement in finetuning for the GEM-Base is almost saturated, which could be a bottleneck to improvement in transfer learning. Overall, as the scale of synthetic data grows, both zero-shot and fine-tuning performances demonstrate a consistent and significant performance improvement.

\begin{table}[]
\centering
\tabcolsep=0.1cm
\caption{The ablation study of the scale of the synthetic data compared with GSD-S.}
\begin{tabular}{llcccccc}
\toprule
Model & Paradigm & Scale  & IoU $\uparrow$    & $F_{\beta} \uparrow$ & MAE $\downarrow$  & BER $\downarrow$   \\
\midrule
&             &  1 $\times$   & 0.703 & 0.819 & 0.215 & 10.79       \\
GEM-T &Zero-Shot    &  5 $\times$      & 0.712 & 0.828 & 0.211 & 11.22         \\
&             &  10 $\times$     & 0.723 & 0.826 & 0.211 & 10.07     \\
&             &  20 $\times$     & 0.726 & 0.844 & 0.206 & 10.98     \\
\midrule
&             &  1 $\times$   & 0.755 & 0.852 & 0.038 & 8.39       \\
GEM-T &Finetune    &  5 $\times$      & 0.757 & 0.855 & 0.035 & 8.54         \\
&             &  10 $\times$     & 0.764 & 0.866 & 0.034 & 8.62        \\
&             &  20 $\times$     & 0.770 & 0.865 & 0.032 & 8.21        \\
\midrule
&             &  1 $\times$   & 0.701 & 0.808 & 0.218 & 11.08       \\
GEM-B &Zero-Shot    &  5 $\times$      & 0.720 & 0.827 & 0.210 & 10.84        \\
&             &  10 $\times$     & 0.725 & 0.830 & 0.209 & 11.20     \\
&             &  20 $\times$     & 0.729 & 0.839 & 0.206 & 10.73     \\
\midrule
&             &  1 $\times$   & 0.766 & 0.873 & 0.031 & 9.44       \\
GEM-B &Finetune    &  5 $\times$      & 0.769 & 0.858 & 0.032 & 8.16         \\
&             &  10 $\times$     & 0.774 & 0.868 & 0.032 & 8.56        \\
&             &  20 $\times$     & 0.774 & 0.865 & 0.029 & 8.35       \\
\bottomrule 
\end{tabular}
\label{tab:transfer111}
\end{table}

\subsection{Ablation Study}
\textbf{Ablation on DQS module.}
The DQS module has influenced the segmentation results in two aspects. One is that it adds an additional binary classification loss from multi-scale features, and the other is that it selects high-quality content features for query initialization in the decoder. To verify the effectiveness of the DQS, we conduct some experiments as shown in Table \ref{tab:ablation}(a). The metric IoU and $F_{\beta}$ decrease by 0.010 and 0.018 when only removing the DQS initialization part, respectively, which proves the improvement of DQS initialization. Surprisingly, we saw a slight increase when removing the extra binary classification loss. We conjecture that the additional loss tends to make the distribution of image features closer to the ground-truth mask, which increases the difficulty of refining randomly initialized queries in the decoder.

\noindent
\textbf{Language prompt in data generation.}
In the data generation process, the language prompt is required as the control inputs, and the multiple prompts can increase the data diversity. We explore the influence of language prompts on the zero-shot and fine-tuning outcomes. In Table \ref{tab:ablation}(b), when utilizing multiple prompts as the language control input, four metrics are consistently superior to the single prompt in the zero-shot setting. As for the fine-tuning stage, the IoU, MAE, and BER metrics of using multiple prompts are better than using a single prompt, and the $F_{\beta}$ metric of these two settings is relatively close. 

\noindent
\textbf{Ablation on the SAM pre-trained model.}
To fully exploit the knowledge of large foundation models, we select the image encoders of MobileSAM and SAM as our backbone. As shown in Table \ref{tab:ablation}(c), the MobileSAM can obtain
non-trivial improvement in terms of four metrics over the ViT-Tiny with ImageNet pre-trained model, e.g., 0.049 IoU and 0.043 $F_{\beta}$ increasement. At the same time, the performance of the ViT-Tiny with ImageNet pre-trained model is obviously higher than the one with random initialization on four metrics, which verifies the importance of image backbone initialization.

\begin{table}[]
\centering
\tabcolsep=0.18cm
\caption{Ablation studies on the GSD-S validation set. }
\begin{tabular}{ccccccccc}
\toprule
\rowcolor{color0}
  &     & IoU $\uparrow$    & $F_{\beta} \uparrow$ & MAE $\downarrow$  & BER $\downarrow$   \\
\hline
\rowcolor{color1}
\multicolumn{6}{l}{(a) Ablation comparison of the proposed DQS.} \\
\hline
\rowcolor{color2}
Extra loss & DQS &&&& \\
\hline
\rowcolor{color2}
 \ding{55} & \ding{55}  & 0.739 & 0.840  & 0.041     & 8.73          \\
 \rowcolor{color2}
\checkmark   & \ding{55}   & 0.734  & 0.831  & 0.043 &  8.80       \\
\rowcolor{color2}
\checkmark            & \checkmark  & 0.744     & 0.849  & 0.043  & 8.92     \\

\hline
\rowcolor{color3}
\multicolumn{6}{l}{(b) Ablation of the language prompt used in the ControlNet.} \\
\hline
\rowcolor{color4}
Paradigm & Prompt  &&&&\\
\hline
\rowcolor{color4}
Zero-Shot &  Single   & 0.690 & 0.803 & 0.217 & 11.72       \\
\rowcolor{color4}
Zero-Shot   & Multiple      & 0.703 & 0.819 & 0.215 & 10.79         \\

\rowcolor{color4}
Finetune        &  Single   & 0.752 & 0.854 & 0.039 & 8.72       \\
\rowcolor{color4}
Finetune    &  Multiple     & 0.755 & 0.852 & 0.038 & 8.39         \\
    
\hline
\rowcolor{color5}
\multicolumn{6}{l}{(c) Comparison of the SAM pre-trained model.}  \\
\hline
\rowcolor{color6}
\multicolumn{2}{c}{Pre-trained model}&&&& \\
\hline
\rowcolor{color6}
\multicolumn{2}{c}{Random}  & 0.551 & 0.681 & 0.091 & 16.90       \\
\rowcolor{color6}
\multicolumn{2}{c}{ImageNet} & 0.695 & 0.806 & 0.048 & 10.97      \\
\rowcolor{color6}
\multicolumn{2}{c}{MobileSAM} & 0.744 & 0.849 & 0.043 & 8.92        \\
\bottomrule
\end{tabular}
\label{tab:ablation}
\end{table}

\subsection{Visualization for Query Selection}
\label{visualization for query selection}
\begin{figure}[t]
  \centering
  \includegraphics[width=3.5in]{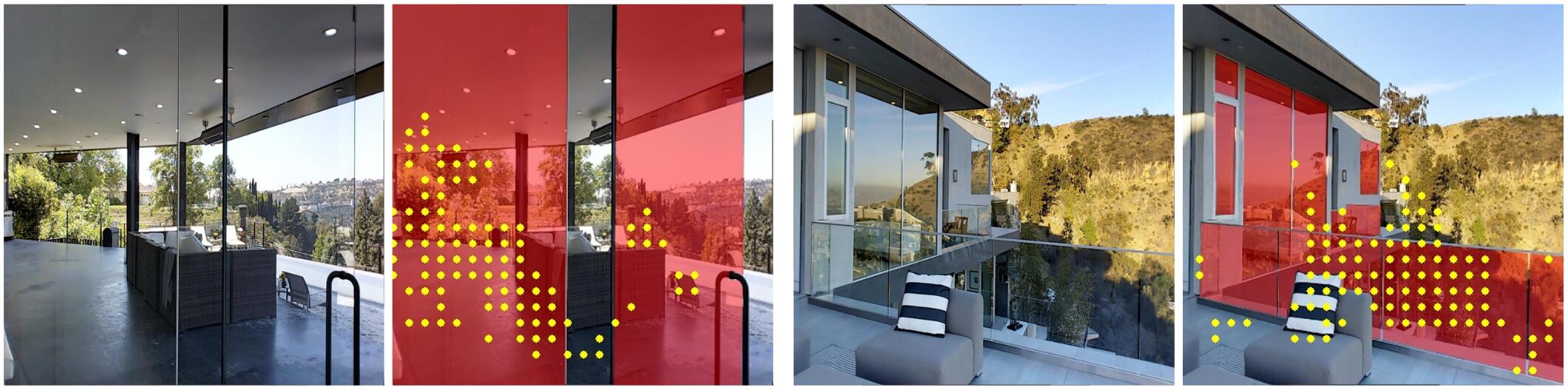}
  \caption{Visualization for discerning query selection. Given two scenarios, the left image depicts the original scene, while the right image highlights the glass mask in red, while the yellow dots indicate selected pixel positions}
  \label{fig:pointvis}
\end{figure}

To validate the effectiveness of our discerning query selection method, we choose two challenging scenarios, as shown in Fig.~\ref{fig:pointvis}. The example aligns with the description provided in Section \ref{discerning query selection}, where the region outside the glass and the glass area collectively constitute the same scene. This configuration poses a significant challenge for the model in distinguishing between the glass and background. Nevertheless, as depicted in Fig.~\ref{fig:pointvis}, the selected queries are predominantly situated within the glass region, resulting in a highly effective initialization of features. Though human observers may hesitate to discern the presence of glass on the left scene, given its extreme transparency and identical content to the background. However, our discerning query selection method successfully designates the left region as the primary selection target, 
which proves the designed query selection method can help our GEM better identify glass regions.

\section{Conclusion}
In this work, we devise a simple glass surface segmentation framework named GEM and propose a synthetic dataset termed S-GSD. Through inserting the image encoder of the SAM model and leveraging synthetic data, the segmentation performance of the simple network is greatly exploited and boosted.
Extensive experiments demonstrate that our GEM achieves a new state-of-the-art on the GSD-S validation set. Besides, we verify that the foundation model can benefit glass segmentation a lot, using the general segmentation model and the diffusion model. We also identify a bottleneck in the improvement when the number of pre-trained data becomes sufficiently large.
Hopefully, this completely new solution could bring inspiration to the study of visual perception integrated with AI-generated content.

\appendix





\bibliographystyle{named}
\bibliography{ijcai24}

\clearpage
\appendix
\begin{center}
    \Large
    Supplimentary Materials
    \normalsize 
\end{center}

\setcounter{section}{0}
\setcounter{table}{0}
\setcounter{figure}{0}
\renewcommand\thesection{\Alph{section}}

\section{Synthetic Data Analysis}
We make a deep comparison and analysis on three publicly available glass surface segmentation datasets with our proposed synthetic dataset, S-GSD, in terms of data scale and data distribution. 

There are three released available glass surface segmentation datasets, including GDD \cite{mei2020don}, GSD \cite{lin2021rich}, and GSD-S \cite{lin2022exploiting}. The GDD dataset is constructed from manual collection and annotation with ground truth glass masks and contains 3916 pairs of samples. 
The GSD dataset consists of two parts, which totally includes 4012 real images with glass surfaces and corresponding masks. One part is collected from existing datasets as well as from the Internet, and the other part is captured by several smartphones. The annotation of GSD is created by manually labeled using Labelme\footnote{https://github.com/labelmeai/labelme}. 
The GSD-S is completely organized from existing semantic segmentation datasets, but the ground truth annotation in this dataset is refined since the glass mask in the original versions is in general inconsistent. This dataset consists of 4519 images. In contrast to the preceding two datasets, the GSD-S dataset contains an extra semantic segmentation mask for the whole scene.

Our synthetic dataset S-GSD consists of four different scales; that is, the number of images is 1x, 5x, 10x, and 20x to the GSD-S training set, respectively. The image of S-GSD is completely generated by the Stable Diffusion \cite{rombach2022high}. To construct the labeled data without any manual annotation, we introduce the ControlNet \cite{zhang2023adding} to the Stable Diffusion for controlling the spatial localization of the synthetic image. Therefore, we can collect numerous images by using the precise mask prior in public available dataset as control conditions. It is worth noting that we only use the mask annotations in GSD-S training set as control conditions for avoiding the leakage of validation information.

\begin{figure}[t]
  \centering
  \includegraphics[width=3.35in]{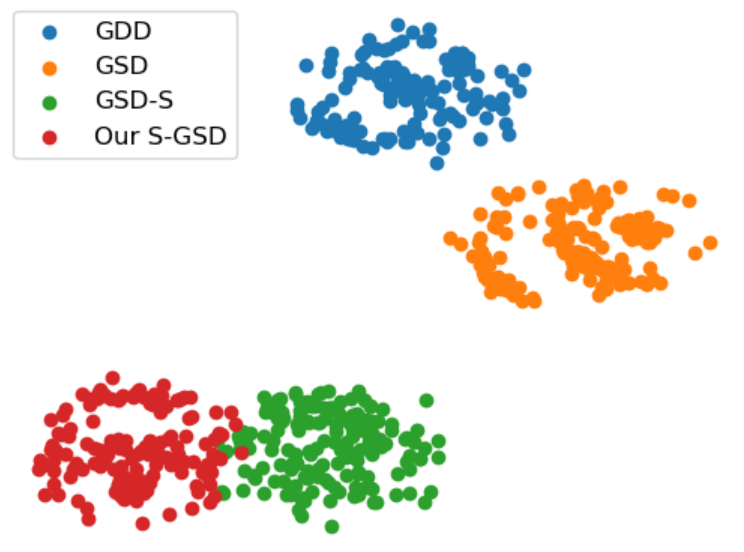}
  \caption{Visualization of feature distribution on our synthetic data and three real datasets utilizing the t-SNE algorithm. The image features are extracted using the CLIP image encoder.}
  \label{fig:point_data_distribution}
\end{figure}

Fig. \ref{fig:point_data_distribution} illustrates the feature distribution of our synthetic data and three real datasets utilizing the t-SNE algorithm \cite{van2008visualizing}. The image features are extracted using the CLIP \cite{radford2021learning} image encoder. The distribution of our synthetic data aligns closely with the GSD-S data distribution, while the distribution of the GDD dataset differs from that of the GSD-S data. This observation is in line with the experimental result that the zero-shot performance of GDD is noticeably lower than that of our S-GSD-1x dataset.

\begin{table*}[]
\centering
\caption{The comparison of different glass surface segmentation datasets.}
\begin{tabular}{l|c|c|cccc}
\toprule
Dataset & Image Source & Mask Source & Number   \\
\midrule
GDD     & Manually Captured & Manually Annotated & 3916  \\
GSD     &  Public datasets \& Manually Captured & Public datasets \& Manually Annotated & 4102  \\
GSD-S   & Public datasets & Manually Relabled & 4519  \\
\midrule
Our S-SGD-1x & \multirow{4}{*}{Auto-generated by Stable Diffusion with ControlNet} & \multirow{4}{*}{Mask prior in GSD-S Training Set} & 3912  \\
Our S-SGD-5x & & & 23467  \\
Our S-SGD-10x & & & 46933  \\
Our S-SGD-20x & & & 93865  \\
\bottomrule
\end{tabular}
\label{tab:data_conparison}
\end{table*}

\section{Synthetic Data Visualization}
\begin{figure*}[!ht]
  \centering
  \includegraphics[width=7.1in]{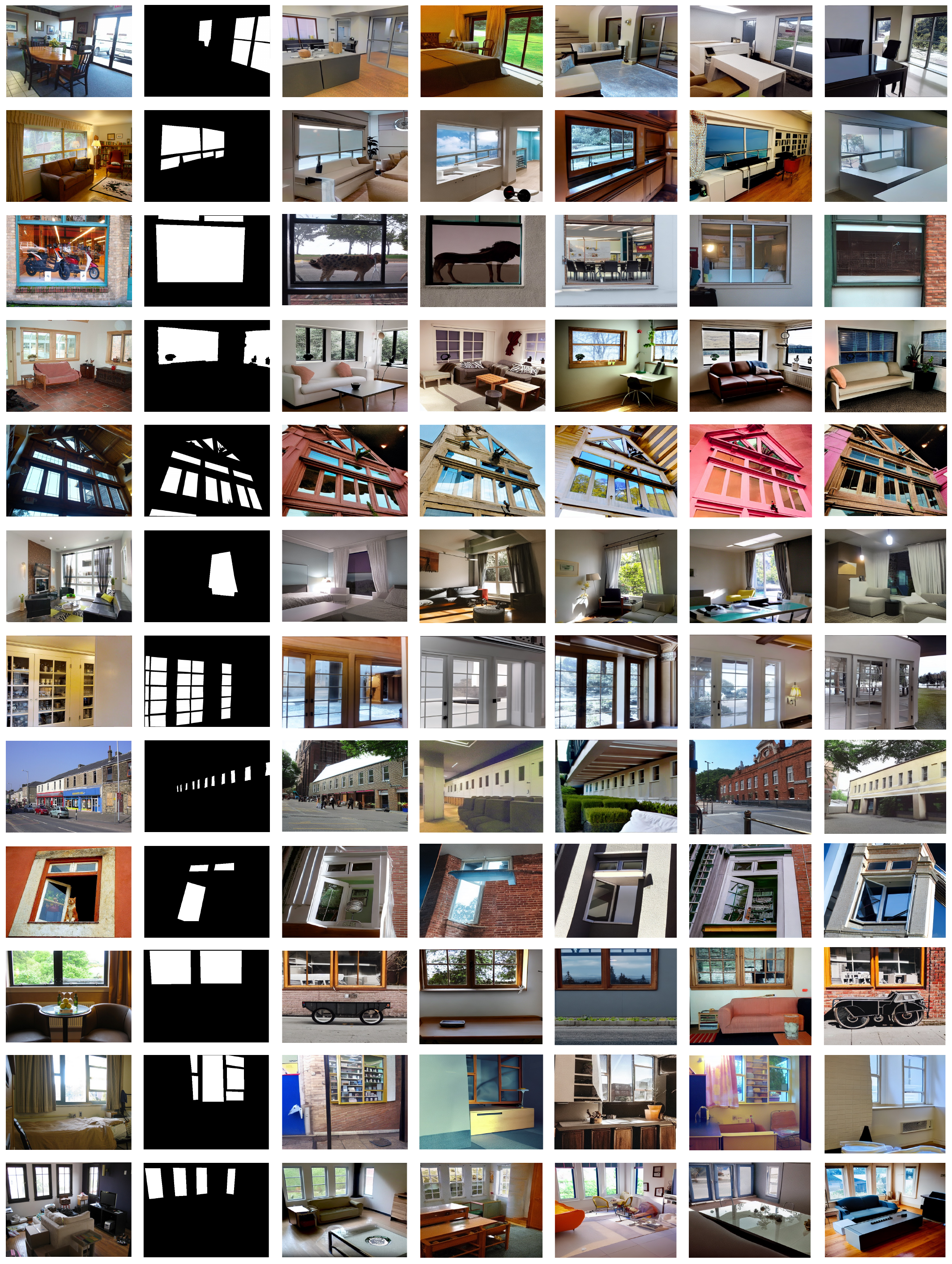}
  \caption{Visual examples of synthetic data. The first and second columns represent the real data and their corresponding masks, respectively. The subsequent columns showcase the synthetic data.}
  \label{fig:crop_synthetic_data}
\end{figure*}

Fig. \ref{fig:crop_synthetic_data} shows more synthetic data. 
The generated image manifests the inherent characteristics of glass, showcasing transparency and reflection. Employing the mask prior as the control condition allows for the creation of training sample pairs comprising the generated images and the corresponding mask. Moreover, when utilizing the same mask prior, the generated images demonstrate a remarkably high level of diversity in terms of the color, weather, background, and object behind the glass.

\section{Training Parameters}
Table \ref{tab:training_parameters} displays the detailed training parameters in terms of GEM-Tiny and GEM-Base. The only difference between these two models is the learning rate. As for the loss in the training stage, we simply follow the Mask DINO \cite{li2023mask}. The total loss is a linear combination of three kinds of losses: $\lambda_{cls}\mathcal{L}_{cls}+\lambda_{L1}\mathcal{L}_{L1}+\lambda_{giou}\mathcal{L}_{giou}+\lambda_{ce}\mathcal{L}_{ce}+\lambda_{dice}\mathcal{L}_{dice}$.

\section{Performance Comparison with other State-of-the-Arts}
Table \ref{tab:GSD-S-Comparisons} illustrates the performance comparison between our GEM model and other state-of-the-art methods. Our GEM exhibits superior glass surface segmentation performance along with commendable efficiency. Specifically, GEM-Tiny and GEM-Base achieve IoU scores of 0.770 and 0.774, respectively, surpassing the performance of GDNet, GlassNet, and GlassSemNet, specially designed for glass segmentation. Importantly, our methods exhibit processing speeds nearly two to three times faster than GlassSemNet \cite{lin2022exploiting} which was the previous state-of-the-art method. This positions our methods as computationally efficient solutions for glass surface segmentation.
Comparison with the latest general segmentation networks, including MaskFormer, Mask2Former, Mask DINO, FASeg, MP-Former, and NAT, reveals that our methods significantly outperform them in terms of all accuracy metrics. While these networks achieve IoU metrics ranging from 0.707 to 0.734, our methods consistently exhibit higher performance.
Besides, the processing speed of our method is still on par with most of its counterparts, except the MaskFormer. However, the metric IoU of our GEM-Base is higher 6.7\% than MaskFormer.

\begin{table}[t]
\centering
\caption{The training parameters for GEM.}
\begin{tabular}{l|c|ccccc}
\toprule
Training Parameters & GEm-Tiny & GEM-Base   \\
\midrule
learning rate  & 2e-4 & 5e-5  \\
weight decay     & 0.05 & 0.05 \\
batch size   & 32 & 32   \\
image size & 384 & 384 \\
epochs for pretrain & 160 & 160  \\
epochs for finetune & 80 & 80  \\
\# queries & 100 & 100  \\
dropout & 0.0 & 0.0 \\
$\lambda_{cls}$ & 4 & 4 \\
$\lambda_{L1}$ & 5 &5 \\
$\lambda_{giou}$ & 2&2 \\
$\lambda_{ce}$ & 5&5 \\
$\lambda_{dice}$ & 5&5 \\
\bottomrule
\end{tabular}
\label{tab:training_parameters}
\end{table}

\begin{table}[t]
\centering
\caption{The evaluation results on GSD-S. Some scores of competing methods are taken from GlassSemNet \protect\cite{lin2022exploiting}.}
\begin{tabular}{>{\small}lcccccccc}
\toprule
Methods      & IoU $\uparrow$    & $F_{\beta} \uparrow$ & MAE $\downarrow$  & BER $\downarrow$   & FPS $\uparrow$ \\
\midrule
SCA-SOD         & 0.558  & 0.689     & 0.087 & 15.03  &  -    \\
SETR            & 0.567  & 0.679     & 0.086 & 13.25  &  -    \\
Segmenter       & 0.536  & 0.645     & 0.101 & 14.02  &  -    \\
Swin            & 0.596  & 0.702     & 0.082 & 11.34  &   -   \\
ViT             & 0.562  & 0.693     & 0.087 & 14.72  &  -    \\
Twins           & 0.590  & 0.703     & 0.084 & 12.43  &  -    \\
SegFormer       & 0.547  & 0.683     & 0.094 & 15.15  &  -    \\
MaskFormer      & 0.707  & 0.826    & 0.043 & 10.91  &  \textbf{28.90} \\
Mask2Former     & 0.732  & 0.838     & 0.043 & 8.93  & 16.20  \\
Mask DINO       & 0.665   &  0.808     & 0.051  & 13.69   &  13.01   \\
FASeg           & 0.725  & 0.843     & 0.048 & 10.26  & 11.32     \\ 
MP-Former       & 0.734  & 0.827     & 0.042 & 8.67  & 16.59    \\
NAT             & 0.730  & 0.846     & 0.041 & 10.16  & 16.68     \\
\midrule
GDNet           & 0.529  & 0.642     & 0.101 & 18.17  & -     \\
GlassNet        & 0.721  & 0.821     & 0.061 & 10.02  & -     \\
GlassSemNet     & 0.753  & 0.860     & 0.035 & 9.26   & 5.64     \\
\midrule
GEM-Tiny        & 0.770 & 0.865 & 0.032 & \textbf{8.21} & 16.09     \\
GEM-Base        & \textbf{0.774} & \textbf{0.865} & \textbf{0.029} & 8.35 & 11.55     \\
\bottomrule
\end{tabular}
\label{tab:GSD-S-Comparisons}
\end{table}

\section{Test on More Real-World Scenes}
\begin{figure*}[t]
  \centering
  \includegraphics[width=6.6in]{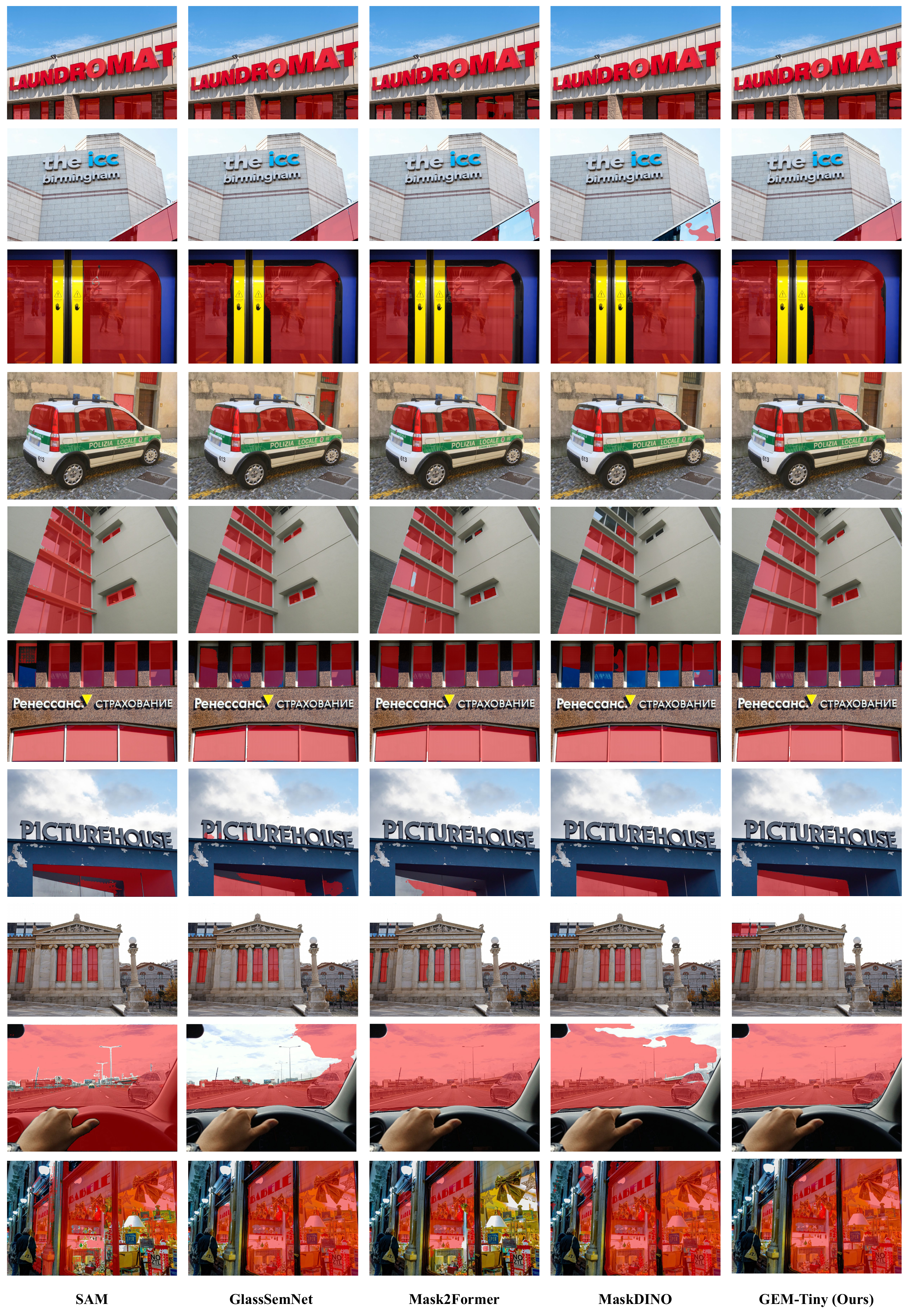}
  \caption{Visualization comparisons of our GEM-Tiny and other state-of-the-art methods tested on real-world scenes.}
  \label{fig:realdata}
\end{figure*}

We also conduct additional tests on real-world scenes, which proves the generalization of our model. The test images are derived from the SA-1B dataset\footnote{https://segment-anything.com/dataset/index.html}, which is released by SAM. This dataset is composed of images licensed from a photo provider that works directly with photographers.
The segmentation results of SAM, GlassSemNet, Mask2Former, MaskDINO, and our GEM-Tiny are presented in Fig.~\ref{fig:realdata}. For SAM, we employ Grounded-SAM\footnote{https://github.com/IDEA-Research/Grounded-Segment-Anything}, utilizing text prompt as the "window" for targeted segmentation. While SAM achieves the best segmentation results in the first four scenes, its performance deteriorates to varying degrees in the last six scenarios. We attribute this decline to SAM's class-agnostic nature. Our GEM-Tiny outperforms the other segmentation methods on these scenes, demonstrating accurate region judgment and superior edge delineation detail. Among the other three segmentation methods, GlassSemNet, Mask2Former, and MaskDINO, each exhibits shortcomings, experiencing failures in specific scenes and often producing incorrect or missing segmentation.

\end{document}